\documentclass[10pt,twocolumn,letterpaper]{article}

\usepackage{wacv}
\usepackage{times}
\usepackage{epsfig}
\usepackage{graphicx}
\usepackage{amsmath}
\usepackage{amssymb}
\usepackage{booktabs}
\usepackage{tabularx}
\usepackage[flushleft]{threeparttable}
\usepackage{multirow}
\usepackage{caption}
\usepackage{fixltx2e}

\newcolumntype{?}{!{\vrule width 1pt}}

\newcolumntype{M}{>{$}c<{$}}
\newcolumntype{C}{>{\centering\arraybackslash}p}
\newcolumntype{Y}{>{\centering\arraybackslash}X}
\def\wacvPaperID{1774} % Enter the WACV Paper ID here

\wacvalgorithmstrack   % Uncomment this line if you are submitting to the Algorithms Track.
\wacvfinalcopy % *** Uncomment this line for the final submission

\ifwacvfinal
\usepackage[breaklinks=true,bookmarks=false]{hyperref}
\else
\usepackage[pagebackref=true,breaklinks=true,colorlinks,bookmarks=false]{hyperref}
\fi

\begin{document}

%%%%%%%%% TITLE
\title{Learning an Ensemble of Deep Fingerprint Representations}

\author{Akash Godbole\\
% Michigan State University\\
{\tt\small godbole1@msu.edu}
\and
Karthik Nandakumar\\
% MBZ University of Artificial Intelligence\\
{\tt\small karthik.nandakumar@mbzuai.ac.ae}
\and
Anil Kumar Jain\\
% Michigan State University\\
{\tt\small jain@cse.msu.edu}
}

\maketitle
\thispagestyle{empty}

%%%%%%%%% ABSTRACT
\begin{abstract}
   Deep neural networks (DNNs) have shown incredible promise in learning fixed-length representations from fingerprints. Since the representation learning is often focused on capturing specific prior knowledge (e.g., minutiae), there is no universal representation that comprehensively encapsulates all the discriminatory information available in a fingerprint. While learning an ensemble of representations can mitigate this problem, two critical challenges need to be addressed: (i) How to extract multiple diverse representations from the same fingerprint image? and (ii) How to optimally exploit these representations during the matching process? In this work, we train multiple instances of DeepPrint (a state-of-the-art DNN-based fingerprint encoder) on different transformations of the input image to generate an ensemble of fingerprint embeddings. We also propose a feature fusion technique that distills these multiple representations into a single embedding, which faithfully captures the diversity present in the ensemble without increasing the computational complexity. The proposed approach has been comprehensively evaluated on five databases containing rolled, plain, and latent fingerprints (NIST SD4, NIST SD14, NIST SD27, NIST SD302, and FVC2004 DB2A) and statistically significant improvements in accuracy have been consistently demonstrated across a range of verification as well as closed- and open-set identification settings. The proposed approach serves as a wrapper capable of improving the accuracy of any DNN-based recognition system.
\end{abstract}

%%%%%%%%% BODY TEXT
\vspace{-1em}
\section{Introduction}

The choice of data representation plays a critical role in determining the success of a machine learning model because different representations can highlight and/or suppress different factors of variation underlying the data \cite{Bengio2013RepLearning}. In fingerprint recognition, domain-specific prior knowledge has played the dominant role in determining the representation scheme, leading to mostly hand-designed features. Since the late $19^{th}$ century \cite{galton1892finger}, it was well-known 
\begin{figure}[ht]
\begin{center}
\includegraphics[width=1.002\linewidth]{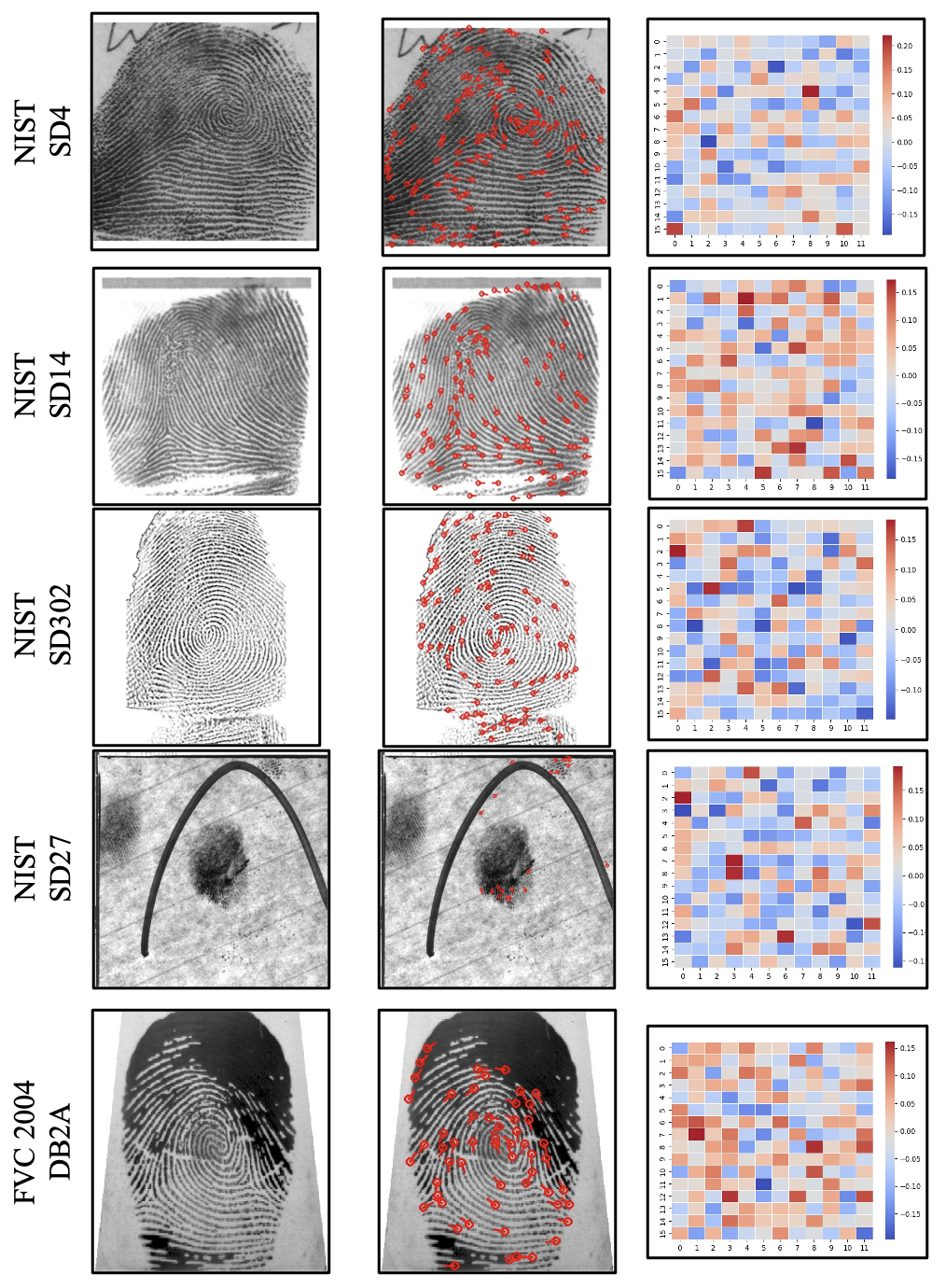}
\caption{Sample fingerprint images, their ISO/IEC representations (minutiae points) \cite{ISO19794}, and heatmap visualizations of 192-dimensional deep neural network representation \cite{engelsma2019learning} from five databases: NIST SD4, NIST SD14, NIST SD302 (N2N), NIST SD27 and FVC 2004 DB2A.}
\label{fig:images}
\end{center}
\vspace{-2em}
\end{figure}
that minutiae are important for identifying fingerprints accurately. Hence, minutiae-based fingerprint representations have become the de-facto standard \cite{ISO19794} as shown in Figure \ref{fig:images}. However, in challenging scenarios such as matching latent fingerprints (see middle column of row 4 in Figure \ref{fig:images}), using only minutiae-based representation is clearly inadequate.

\begin{figure*}[ht]
\begin{center}
\includegraphics[width=0.9\linewidth]{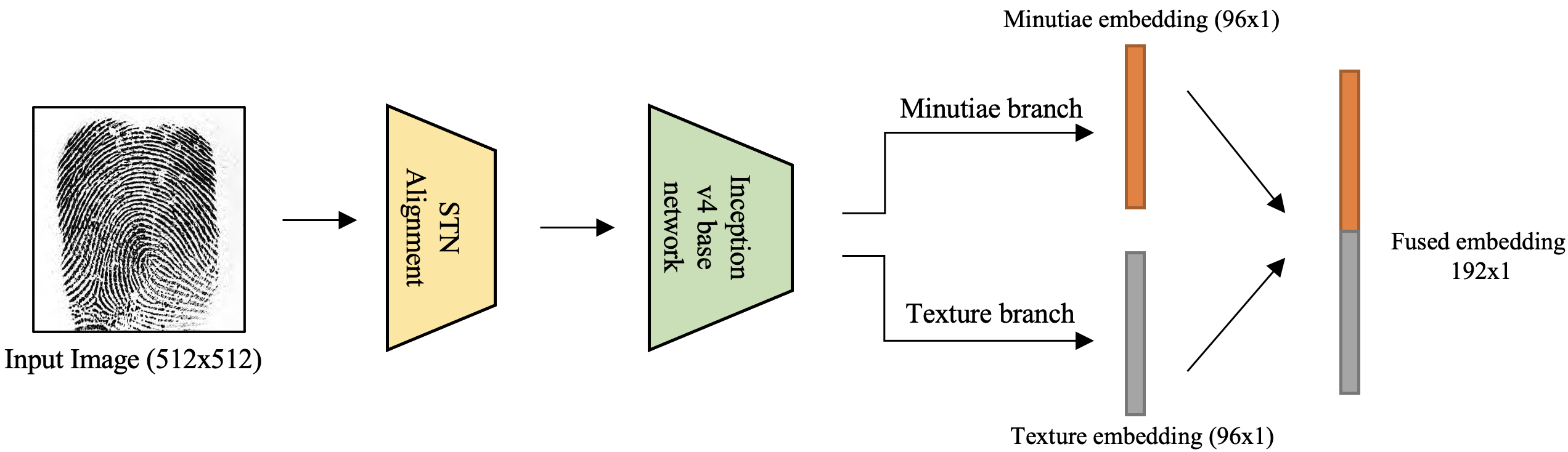}
\caption{An overview of the DeepPrint architecture \cite{engelsma2019learning}. STN stands for Spatial Transformer Network.} 
\label{fig:dp_overview}
\end{center}
\vspace{-1.5em}
\end{figure*}

With the advent of deep learning and its tremendous success in various applications including consumer sentiment analysis \cite{jain2021hybrid, yadav2020sentiment}, biometric recognition \cite{sundararajan2018deep}, natural language processing (NLP) \cite{deng2018deep}, healthcare, and finance \cite{heaton2017deep}, the concept of data-driven representation learning has come to the fore. It is possible to learn multiple representations from the same data by applying different priors, which are usually determined by the architecture (and depth) of the neural network, ground-truth labels that guide/supervise the learning, and the objective/loss function. It is well-known that no single prior can perfectly disentangle all the underlying variations in data and lead to a universally good representation. Consequently, the idea of \emph{ensemble learning} \cite{he2016deep, seni2010ensemble}, which refers to learning multiple models/representations (as opposed to using a single representation) has been used to improve the diversity of the feature space. This approach has been successfully employed in many computer vision tasks to boost performance compared to a single model \cite{sagi2018ensemble}.

\par In the field of biometrics, many studies have been conducted over a wide range of modalities (face, fingerprint, gait, lip, etc.) using different implementations of ensemble learning \cite{best2014unconstrained, paulino2010latent, marcialis2005fusion, marcialis2004fingerprint, basheer2021fesd, porwik2019ensemble, gupta2020multiple}. While some of these methods may not fit into the traditional definition of ensemble learning, they can be considered to be a part of this family since they involve some form of fusion of outputs obtained from different entities. There are two key challenges involved in implementing any ensemble learning approach: (i) generation of multiple representations from the same input that are sufficiently discriminative as well as diverse, and (ii) efficient fusion of these representations during the inference process. Typically, the first problem is solved by learning multiple representations based on different augmentations of the training data, different network architectures, or different data partitions \cite{zhong2016overview}. The latter issue is addressed through a range of early (feature-level) and late (score- or decision-level) fusion techniques \cite{ganaie2021ensemble}.

In this work, our objective is to improve the performance of a state-of-the-art (SOTA) deep neural network based fingerprint recognition model called DeepPrint (DP) \cite{engelsma2019learning} through ensemble learning. The core advantage of the DP model is its ability to learn a compact fingerprint representation using a combination of domain knowledge (minutiae features) and data-driven (texture patterns) supervision techniques. Figure \ref{fig:dp_overview} presents an overview of the DP architecture and the third column of Figure \ref{fig:images} shows heatmap visualizations of the DP representation. Despite its strong ability to extract discriminative information from fingerprint images, the performance of DP models still fall short of commercial-off-the-shelf (COTS) fingerprint recognition systems (which use both minutiae and other proprietary features) in challenging scenarios. We posit that this is primarily due to the reliance on a single representation, which fails to capture all useful information. To overcome this limitation, we make the following contributions in this work:

\begin{figure*}[t]
\begin{center}
\includegraphics[width=\linewidth]{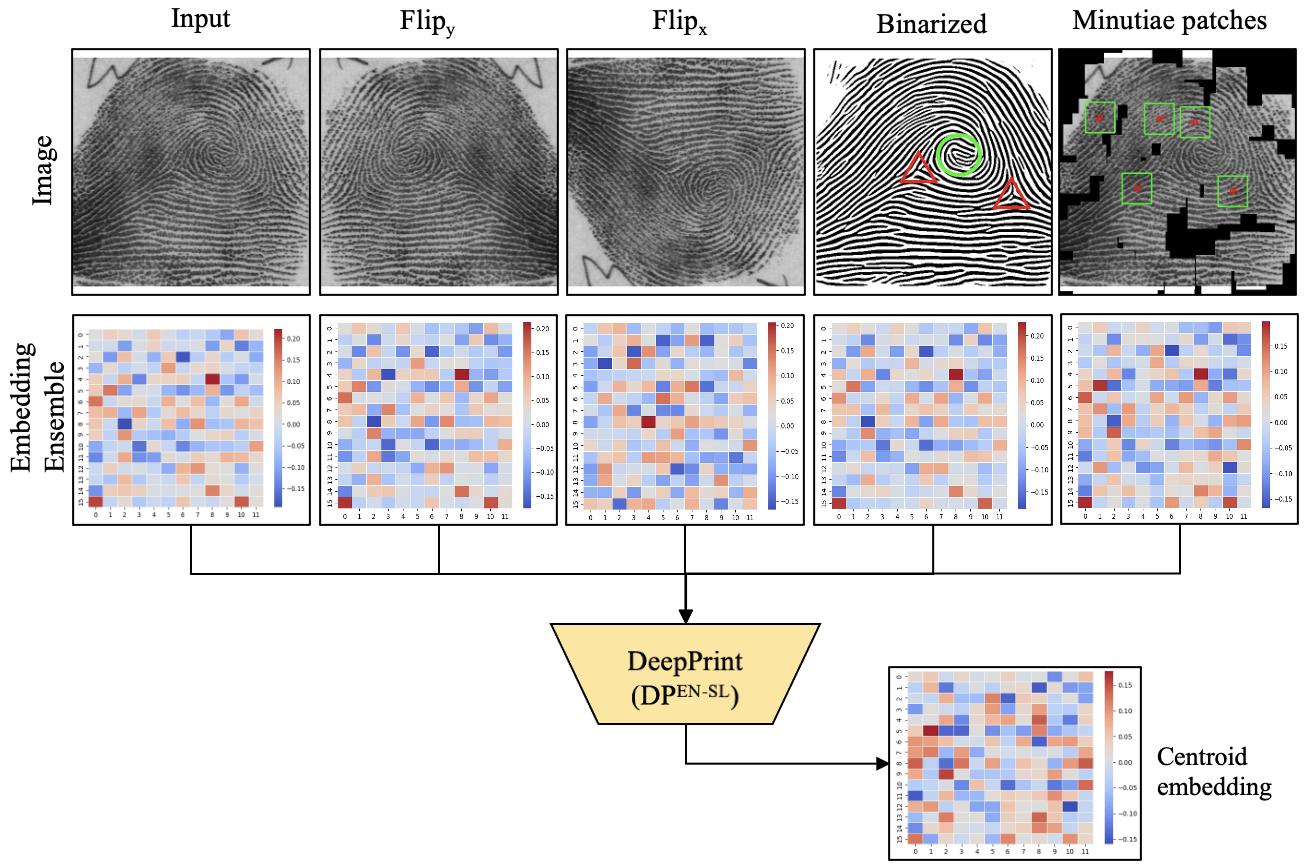}
\caption{Four different image manipulations performed on the original image. \textit{Flip\textsubscript{x}} and \textit{Flip\textsubscript{y}} were obtained using OpenCV \cite{2015opencv}. The binarized ridge (flow) image and minutiae locations were obtained using Verifinger 12.3 SDK. For illustration purposes, the singular points of the fingerprint are highlighted in the ridge flow image: core (green circle); delta (red triangles), and only a few minutiae patches are shown in the soft-gated minutiae image, where the blurred pixels are set to 0 for better visual clarity. The representations shown in the second row are extracted from the model trained on the corresponding image types in the top row. The bottom row shows the feature fusion strategy used in this study.}
\label{fig:ensemble_images}
\end{center}
\vspace{-1.5em}
\end{figure*}

\vspace{-0.5em}
\begin{itemize}
    \item Generating an ensemble of five fingerprint representations from a single image using the DeepPrint architecture. This is achieved by augmenting the original training data with four types of image manipulations (two generic and two domain-specific transformations).

    \item Evaluation of the DeepPrint-based ensemble of fingerprint embeddings using decision and score fusion schemes on five fingerprint databases - NIST SD4, NIST SD14, FVC 2004 DB2A, NIST SD27, and NIST SD302 (N2N) \cite{sd4, sd14, fvc, sd27, nist302}. While this approach improves accuracy, it comes at the cost of increased computational complexity (and lower throughput).

    \item Training a single DeepPrint model on unperturbed images, which is capable of learning the diversity present in the ensemble of fingerprint representations. This distilled model can be considered as a feature fusion strategy that learns a single embedding through external supervision at the feature level from the component representations in the ensemble. 

    \item Comprehensive experiments in the verification and identification (both closed and open-set) modes to demonstrate that the proposed feature fusion approach can consistently improve the accuracy \emph{without compromising on retrieval or feature extraction times}.
   
\end{itemize}

\section{Related Work}

Ensemble learning has proven to be an effective tool for improving the generalization ability of deep learning models \cite{dong2020survey, sagi2018ensemble}. Moreover, many classical machine learning algorithms such as AdaBoost \cite{freund1996experiments}, Bagging \cite{breiman1996bagging}, and Random forest \cite{breiman2001random, amit1994randomized} also use ensemble learning at their core. Broadly, ensemble learning can be divided into two stages.

% Figure \ref{fig:schematic} shows a schematic diagram illustrating the main components of ensemble learning. 

\subsection{Generation of Ensemble}

There are several methods of generating an ensemble of features/models that have varying degrees of complexity: 

\vspace{-1em}

\begin{itemize}
\item \textbf{Altering the training data}: This method manipulates the training dataset in various ways and trains one instance of the network architecture for each altered training set independently. Manipulation methods may range from common data augmentation techniques like contrast adjustment and rotation to more complex manipulations like binarization and gradient images \cite{sagi2018ensemble, manipulations}. This is the approach used in this study for generating the ensemble of fingerprint embeddings. 
\vspace{-1em}

\item \textbf{Altering the architecture of the network}: In this method, the network architecture is changed before training each model in the ensemble. These changes could be applied to hyperparameter configurations of the training process or manipulating the overall design of the network itself. This method is considered more complex since it is difficult to ensure that the changes made in the architecture do not dramatically affect the individual accuracy of each model in the ensemble. Another variation of this concept is using entirely different networks as part of the ensemble \cite{netensemble,simonyan2014very,he2016deep}.
\vspace{-1em}
\item \textbf{Data partitioning}: This method splits the training dataset into smaller subsets and trains a given network architecture on each subset, generating an ensemble of models that are trained on subsets of the original training set. The partitioning of the training datasets can be done at a sample-level or at a feature-level \cite{piao2015new, breiman1996bagging, fred2005combining}. Sample-level partition reduces the size of the training dataset for each model and the resulting models have limited diversity. In contrast, feature-level partition can generate diverse representations, but requires careful feature selection techniques to ensure that the resulting models have good accuracy.
\end{itemize}

\vspace{-1em}

\subsection{Information Fusion for Ensemble Learning}

\par The second step in ensemble learning is the fusion of outputs generated by individual models in the ensemble \cite{huang2009research}. Some of the common fusion techniques include:

\begin{itemize}
    \item \textbf{Feature fusion}: This method aims to combine the multiple feature representations into a single embedding. The most simple approach is feature concatenation, where the embeddings generated by the ensemble for a given sample are concatenated to yield a single, large embedding. An alternative approach that maintains the original feature dimensionality is knowledge distillation \cite{hinton2015distilling}, where a `student network' is learned through external supervision from the `teacher networks' that constitute the ensemble.
    
    \item \textbf{Score fusion}: This technique combines the prediction confidences of the individual models into a single value, which is considered as the output of the ensemble. This is typically achieved by computing the sum, mean, median, weighted sum, or weighted mean of the individual outputs \cite{dong2020survey}. 

    \item \textbf{Decision fusion}: In this method, the final prediction of the ensemble is determined by combining the predictions of the individual models. The simplest case is a majority voting scheme, where the decision favored by a majority of the models in an ensemble is considered as the final output \cite{dong2020survey}. In the case of identification, it is also possible to combine ranks output by models in the ensemble leading to rank fusion schemes \cite{Kumar2009}.

\end{itemize}

\begin{table*}[h]
    \centering
    \captionsetup{justification=centering}
    \caption{Properties of the 5 databases used in this study.}
    \begin{threeparttable}
    
    \begin{tabular}{|c|C{0.15\linewidth}|C{0.2\linewidth}|c|c|}
    \hline
         \textbf{Database} & \textbf{\# of identities (unique fingers)}  &\textbf{\# of images} & \textbf{Type of fingerprints} & \textbf{Availability}\\
        %  \hline
        %  & & $\nu=1$ & $\nu=2$ & $\nu=3$ & Mean score-fusion & Median score-fusion & \\
         \hline
         NIST SD4 \cite{sd4} \tnote{*} & 2,000 & 4,000 (2/finger) & Rolled & Private\\
         \hline
         NIST SD14 \cite{sd14}\tnote{*}& 27,000 & 54,000 (2/finger) & Rolled & Private\\
         \hline
         FVC2004 DB2A \cite{fvc}& 100 & 800 (8/finger) & Plain (distorted) & Public \\
         \hline
         NIST SD27 \cite{sd27}\tnote{*}& 258 & 516 (latent and mated rolled) & Latent, Rolled & Private \\
         \hline
         NIST SD302 (N2N) \cite{nist302} \tnote{†}& 2,000 & 25,093 (10-15/finger) & Rolled, Plain & Public \\
         \hline
        %  NIST SD302 (N2N-RL) \cite{nist302} \tnote{†}& 1,019 & 2,774 (2-10/finger) & Rolled, Latent & Public \\
        %  \hline
    \end{tabular}
    \begin{tablenotes}
        \item[*] NIST SD4, NIST SD14, and NIST SD27 have been retracted from the public domain due to privacy concerns.
        % \item[†] NIST SD302 consists of multiple subsets containing different types of fingerprints. N2N-RP subset contains only Rolled and Plain fingerprints. N2N-RL subset contains Latent fingerprints and their exemplar rolled fingerprints.
        \item[†] NIST SD302 consists of multiple subsets containing different types of fingerprints. The subset used in this study contains only Rolled and Plain fingerprints.
    \end{tablenotes}
    \end{threeparttable}
    \label{tab:database_stats}
\end{table*}

% \begin{figure}[t]
% \begin{center}
% \includegraphics[width=1.002\linewidth]{images/databases.png}
% \caption{Four images from each of the five databases that are used for evaluation of the ensemble of fingerprint embeddings in this study.}
% \label{fig:databases}
% \end{center}
% \vspace{-1.5em}
% \end{figure}

\begin{figure}[t]
\begin{center}
\includegraphics[scale=0.35]{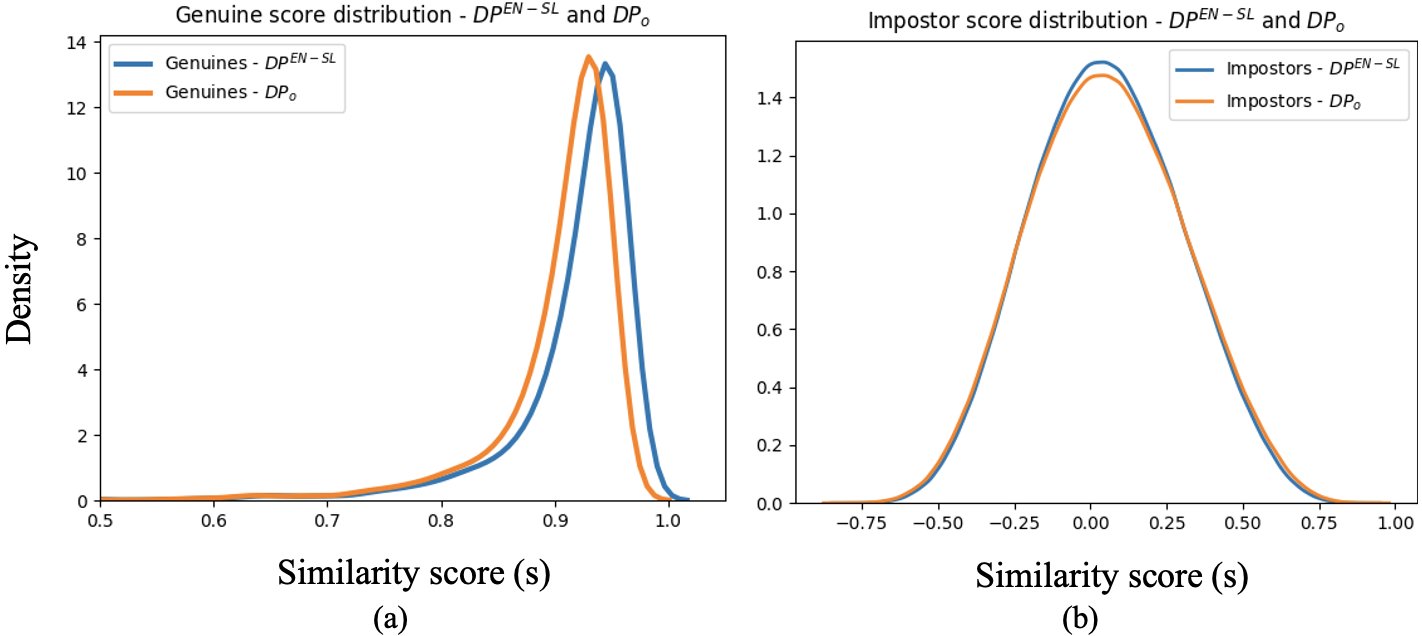}
\caption{Genuine (a) and impostor (b) distributions of DP\textsubscript{o} (baseline) and DP\textsuperscript{EN-SL} (proposed) on NIST SD4. While genuine scores generated by DP\textsuperscript{EN-SL} are higher, impostor distribution of DP\textsuperscript{EN-SL} is more peaked than that of DP\textsubscript{o}.}
\label{fig:gen_imp_dist}
\end{center}
\vspace{-1.5em}
\end{figure}

\vspace{-1em}
\section{Proposed Approach}

\subsection{Fingerprint Ensemble Generation}

The keys to the success of ensemble learning are ``accuracy and diversity'' of the feature representations included in the ensemble \cite{Zhou2021}. Diversity is essential to incorporate as much complementary information as possible within the ensemble, which ensures that the failure modes of the different representations do not overlap significantly. On the other hand, it is important that each representation in the ensemble is accurate and comparable to one another. Lack of either diversity or accuracy can degrade the performance of the ensemble instead of enhancing the accuracy achieved using the best individual model. 

Given the complexity of designing a DNN architecture that works well for fingerprints and limited size of the training datasets, we generate the ensemble using the input manipulation approach. Let the original training set be denoted as $\mathcal{D}_o$. Each fingerprint image in the original training set is perturbed using four manipulation techniques (denoted as \textit{Flip\textsubscript{y}}, \textit{Flip\textsubscript{x}}, \textit{Ridge}, and \textit{Minu}) to obtain four manipulated datasets (denoted as $\mathcal{D}_y$, $\mathcal{D}_x$, $\mathcal{D}_r$, and $\mathcal{D}_m$, respectively) with the same size as $\mathcal{D}_o$. Figure \ref{fig:ensemble_images} shows the manipulations selected in this work, which includes two generic geometric transformations and two transformations that are specific to the fingerprint domain. The two geometric transformations are \textit{Flip\textsubscript{y}} and \textit{Flip\textsubscript{x}}, where the original images are flipped along the $y$ and $x$ axes, respectively. Since convolutional neural network (CNN) architectures such as DeepPrint are not rotation-invariant, \textit{Flip\textsubscript{y}} and \textit{Flip\textsubscript{x}} provide a simple way of constructing additional datasets with geometric operations, while ensuring diversity of generated representations.

%These five representations are denoted as $f_o$, $f_x$, $f_y$, $f_r$, and $f_m$, where $f_* = DP_*(I_*)$ and $DP_*$ is the DeepPrint model obtained through training on dataset $D_*$, $* \in \{o, x, y, r, m\}$.
%  (i.e., $I_y$ = \textit{Flip\textsubscript{y}}$(I_o)$ and $I_x$ = \textit{Flip\textsubscript{x}}$(I_o)$).
%($I_r$ = \textit{Ridge}$(I_o)$)
%($I_m$ = \textit{Minu}$(I_o)$)

Next, we apply the \textit{Ridge} transformation, which is the binarized ridge image extracted using the Verifinger SDK\footnote{https://www.neurotechnology.com/verifinger.html}. Since the binarized ridge images emphasize level-1 (global) features in a fingerprint, including better clarity on the core and delta points, the representation obtained through learning on these images is expected to focus more on the global features. Finally, we create a soft-gated minutiae image, where the \textit{Minu} transformation de-emphasizes the regions of the fingerprint where no minutiae points are detected by the Verifinger SDK. This is achieved by retaining the $64 \times 64$ pixel patches centered at the location of each detected minutia point and applying a Gaussian blur ($k=11$) on the regions of the fingerprint image that are not included in any minutia patch. The representation learned from these soft-gated minutiae images can be expected to further emphasize the level-2 (more local, keypoint) features of a fingerprint.

An instance of the DeepPrint (DP) architecture is trained on each of the above five datasets ($\mathcal{D}_o$, $\mathcal{D}_y$, $\mathcal{D}_x$, $\mathcal{D}_r$, and $\mathcal{D}_m$), resulting in an ensemble of five models denoted as $DP_o$, $DP_y$, $DP_x$, $DP_r$, and $DP_m$, respectively. Our baseline model ($DP_o$) actually performs better than the model originally proposed in \cite{engelsma2019learning} due to some hyperparameter tuning on our part (lowering the learning rate multiplier of the STN). Therefore, in all of our models in the ensemble, we employ this new set of hyperparameters.
% Since hyperparameter tuning of the training procedure introduced in \cite{engelsma2019learning} resulted in a higher base accuracy, all the above DP models were learned using the same optimized hyperparameters.

% \begin{figure*}[t]
% \begin{center}
% \includegraphics[width=\linewidth]{images/schematic.png}
% \caption{Different components of ensemble learning employed in this study.}
% \label{fig:schematic}
% \end{center}
% \vspace{-1.5em}
% \end{figure*}

% \begin{figure*}[t]
% \begin{center}
% \includegraphics[width=\linewidth]{images/tsne_.png}
% \caption{2-dimensional t-SNE plots showing 5 failure cases (image pairs) of DP\textsuperscript{*} and DP\textsuperscript{EN-SL}. The plot on the left shows embeddings from DP\textsuperscript{EN-SL} for 5 image pairs that were misclassified by DP\textsuperscript{*}. The embeddings for the same 5 pairs from DP\textsuperscript{*} are shown in the figure on the right. We can see that the separation between the embeddings is much smaller in DP\textsuperscript{EN-SL} compared to DP\textsuperscript{*} (see axis range for better clarity).}
% \label{fig:tsne_}
% \end{center}
% \vspace{-1.5em}
% \end{figure*}

\subsection{Ensemble Fusion}
\label{sec:output_fusion}

Both early (feature level) and late (score and decision level) fusion schemes are considered in this work. The advantage of late fusion schemes is that no additional training is usually required and it is possible to make full use of the available representations in the ensemble, leading to higher accuracy. The drawback of late fusion techniques is that when comparing two fingerprint images, the ensemble of representations must be generated for \textit{both} the fingerprint images. In the case of identification, the ensemble of representations has to be generated for the \textit{entire} gallery. This can be expected to increase the feature extraction and matching times, thereby reducing the system throughput.

In this study, we use the OR rule for decision-level fusion, which accepts a pair of fingerprint images as a match if at least one of the models in the ensemble outputs a match decision. Note that similarity between a fingerprint pair is computed based on representations of the two images generated using the same model only. This is because each model has its own unique threshold for a given $FMR$ during the training stage and cross-representation similarity scores cannot be interpreted fairly using multiple thresholds. Furthermore, it was observed that the genuine score distribution is disproportionately affected by cross-representation comparisons, while the impostor score distribution remains relatively unaffected, thereby drastically reducing the $TAR$ at a given $FMR$. In the subsequent discussion, the results from decision fusion are denoted as $DP^{EN-DF}$. Mean and median fusion rules are employed for score level fusion. In \textit{mean (median) score fusion}, the mean (median) of similarity scores obtained from each model in the ensemble for a given pair of fingerprints is computed. We only report the results for median score fusion, since it consistently outperformed mean score fusion in all our experiments. Henceforth, score fusion results are referred to as $DP^{EN-SF}$.

Finally, we implement a feature-fusion technique to generate a single embedding (from a single model) that attempts to encapsulate the diverse information contained within the multiple representations. Using every model in the ensemble (i.e., $DP_o$, $DP_{y}$, $DP_{x}$, $DP_{r}$, and $DP_{m}$), we extract the features from each image in the corresponding training set ($\mathcal{D}_o$, $\mathcal{D}_y$, $\mathcal{D}_x$, $\mathcal{D}_r$, and $\mathcal{D}_m$, respectively). We then use all of these five representations (or a subset of them) as external supervisors to train a new DeepPrint model. The resulting model, denoted as $DP^{EN-SL}$, is trained in the same way as $DP_o$ using images from $\mathcal{D}_{o}$, but utilizes the aforementioned supervisors to minimize an additional objective function:

\vspace{-8pt}
\begin{equation}
\label{eqn:mse}
    \mathcal{L}\left(F_e, \{F_c\}_{c \in \mathcal{M}}\right) = \sum_{c \in \mathcal{M}}(\sigma_c * ||F_e - F_c||_2^2),
\end{equation}
\vspace{-8pt}

\begin{table*}
    \centering
    \captionsetup{justification=centering}
    \caption{Verification accuracy of various models on the five fingerprint databases. The values in the table represent true accept rate $TAR(\%)$ at a false match rate $FMR=0.01\%$, except for FVC20024, where TAR at FMR=0.1\% has been reported using 4,950 impostor pairs \cite{fvc}. Note that $DP$ and the five base models in the ensemble have been fine-tuned on a separate latent database before they are evaluated on NIST SD27. Moreover, there is no segmentation or enhancement applied to the latent images in our approach, unlike Verifinger.}
    \begin{threeparttable}
    \begin{tabular}{|c || c | C{0.1\linewidth} | c| c| c|} 
         \hline
         Model & \multicolumn{5}{c?}{\textbf{$TAR (\%)$}}  \\ [1.0ex]
         \noalign{\hrule height 1.2pt}
         & NIST SD4 & NIST SD14 & FVC2004 DB2A & NIST SD27 & N2N\\
         \noalign{\hrule height 1.2pt}
         $DP$ \cite{engelsma2019learning}  &  97.9 & 98.55 & 87.9 & 25.2 & 71.6\\  
         \hline
         $DP_o$  &  98.5 & 98.75 & 88.78 & 26.36 & 74.8\\  
         \hline
         $DP_{x}$  & 98.5 & 98.7 & 88.78 & 26.36 & 74.8\\  
         \hline
         $DP_{y}$  & 98.55  & 98.75 & 88.78 & 26.36 & 74.8\\ 
        \hline
        $DP_{r}$  & 98.5 & 98.75 & 88.67 & 25.6 & 73.45\\  
        \hline
        $DP_{m}$  & 98.45 & 98.7 & 88.57 & 25.2 & 73.3 \\
        \hline
        $DP^{EN-DF}$ & 99.3 & 99.25 & 93.21 & 29.1 & 78.48\\
        \hline
        $DP^{EN-SF}$ & 98.65 & 98.75 & 91.07 & 28.3 & 74.7\\
        \hline
        $DP^{EN-SL}$ & 99.2 & 99.35 & 92.92 & 28.68 & 77.96\\
        \hline
        Verifinger & 99.7 & 99.9 & 96.0 & 46.12 & 84.6 \\
        %  \hline
        %  $M_{binary}$  & 98.5 & 98.65 & 91.5 \\
        %  \hline
        %  $M_{grad}$  & 98.45 & 98.65 & 91.0 \\
         \noalign{\hrule height 1.2pt}
    \end{tabular}
    \end{threeparttable}
    \label{tab:one_to_one}
\end{table*}

\noindent where $F_e$ is the new representation that is generated by the supervised model $DP^{EN-SL}$, $F_{c}$ is the pre-extracted feature representation from model $DP_c$, $\sigma_c$ is a scalar weight assigned to model $DP_c$ in proportion to the accuracy of $DP_c$ relative to the other models in the ensemble, and $\mathcal{M} \subseteq \{o, y, x, r, m\}$. Note that if we ignore the weights, the loss in the above equation is minimized when $F_e$ is the ``centroid'' of the multiple feature representations, which is known to be quite effective in image retrieval tasks \cite{wieczorek2021unreasonable}. However, in contrast to existing techniques that require extraction of multiple embeddings and computation of the centroid at inference time, the proposed $DP^{EN-SL}$ model directly learns to extract the centroid representation from the original image during training. This generates a representation that is more discriminative compared to any of the individual models $DP_c$, thereby improving recognition accuracy. Since there is no need to perturb the given image pair during inference, it has the same throughput as the vanilla DP model. Additionally, the size of the gallery remains unchanged as opposed to other late fusion schemes. Thus, \emph{the proposed feature fusion method enhances accuracy without increasing computational or memory requirements during recognition}, albeit at a higher training cost.

Unless specified otherwise, the models used to supervise $DP^{EN-SL}$ are $DP_{r}$ and $DP_{m}$, since supervision based on these two models yielded the highest accuracy. Moreover, since the baseline accuracy of $DP_{r}$ is generally higher than than that of $DP_{m}$, we assign $\sigma_{r}>\sigma_{m}$ (0.08 and 0.05, respectively). Figure \ref{fig:gen_imp_dist} shows the genuine and impostor distribution on NIST SD4 obtained using $DP_o$ and $DP^{EN-SL}$. We can see that the genuine scores generated by the latter are higher than those of the former. 
% Figure \ref{fig:tsneß} shows the 3-D t-SNE plot of embeddings generated by $DP_o$ and $DP_e$. A close inspection of this plot makes it evident that the embeddings generated by $DP_e$ cover a larger feature space compared to $DP_o$. This suggests that the features obtained from $DP_{e}$ are indeed more diverse than those obtained from $DP_o$, even though both these models are trained on the same input images from $\mathcal{D}_o$. Henceforth, feature fusion results based on the external supervision method described above are referred to as $DP^{EN-SL}$.

\section{Experimental Results}

\subsection{Databases}

\par We consider five fingerprint databases consisting of rolled, plain and latent prints - NIST SD4 (Rolled), NIST SD14 (Rolled), NIST SD27 (Latent-Rolled), N2N (Rolled, Plain) and FVC 2004 DB2A (Plain) for evaluating the proposed methods. 
% N2N-RP and N2N-RL are different subsets of the same database (NIST SD302) and this nomenclature is unique to this study. 
% While N2N-RL consists of latent fingerprints and their mated rolled prints, N2N-RP consists of only rolled and plain prints. 
Table \ref{tab:database_stats} reports the key information about each database. Though NIST SD4 and SD14 have been widely used to evaluate SOTA algorithms in the past, they are no longer available in the public domain. NIST SD14 contains 27,000 fingerprint pairs, but we restrict our evaluation only to the last 2,700 pairs in order to ensure comparability in accuracy with existing studies in the literature.

\subsection{Verification}

\par Table \ref{tab:one_to_one} shows the verification accuracy for the various models on the five evaluation databases used in this study. Some of the keys observations from this table are as follows. Despite being a SOTA DNN method, there is gap between the accuracy of the original DeepPrint model ($DP$) \cite{engelsma2019learning} and the COTS Verifinger matcher, especially when matching the more challenging latent fingerprints (NIST SD27 and N2N-RL datasets). Bridging this gap is the primary motivation for this study. Hyperparameter tuning improves the performance of the original $DP$ model, which explains the difference between the $DP$ \cite{engelsma2019learning} and $DP_o$ models. 

% \begin{figure}[h]
% \begin{center}
% \includegraphics[width=1.009\linewidth]{images/supervised.png}
% \caption{External supervision of the DeepPrint network using pre-extracted representations from the ridge flow ($DP_{r}$) and minutiae patch ($DP_{m}$) models. Only some minutia points are highlighted in the minutiae patch image.}
% \label{fig:supervised}
% \end{center}
% \vspace{-1.5em}
% \end{figure}

\begin{figure*}[h]
\begin{center}
\includegraphics[width=\linewidth]{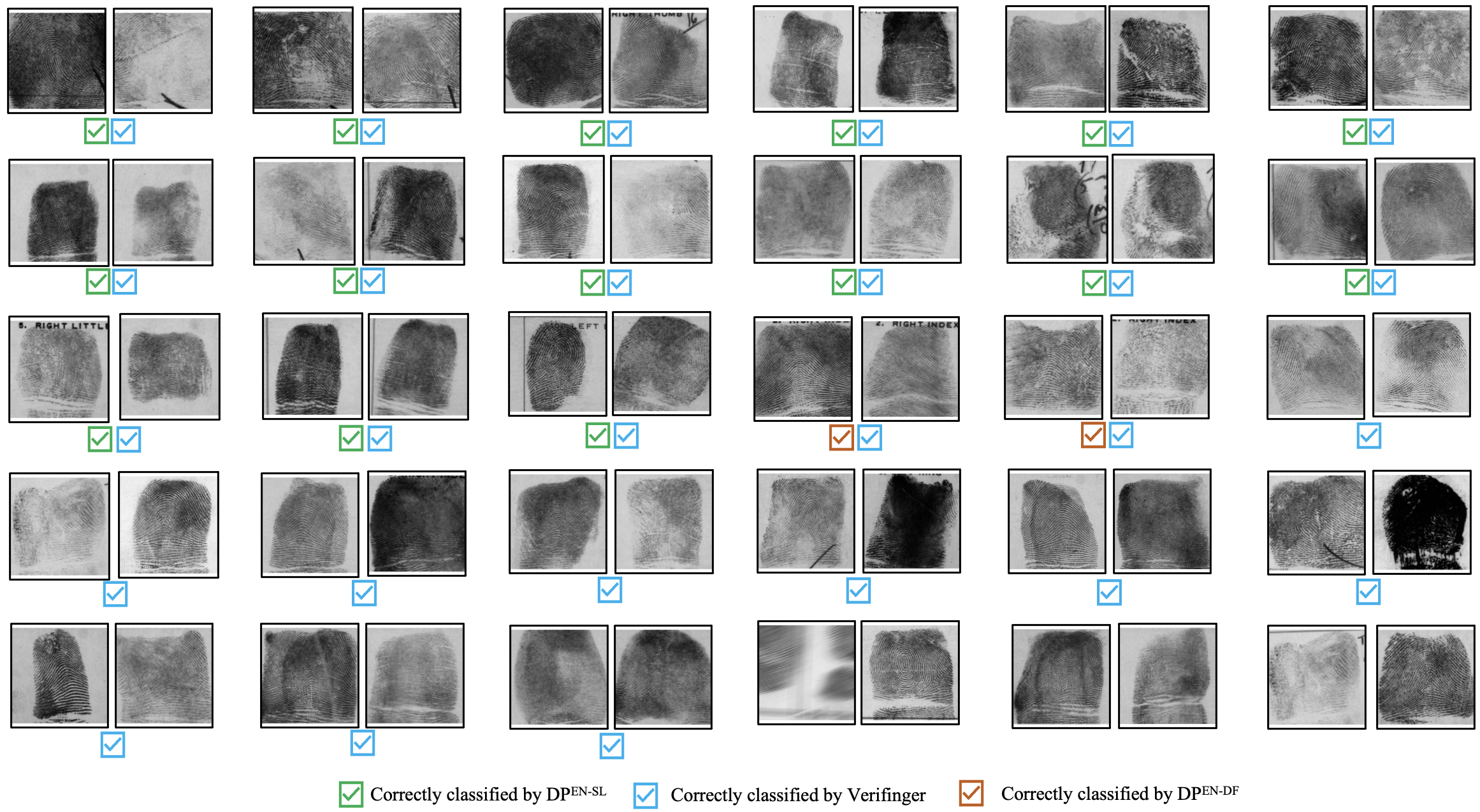}
\caption{30 genuine failure cases reported by DP\textsubscript{o} on NIST SD4. Under each pair of images, there is a indicator that shows whether that pair was correctly classified by DP\textsuperscript{EN-SL}, Verifinger or  DP\textsuperscript{EN-DF} used in this study. The absence of a check mark under a pair of images indicates that neither DP\textsuperscript{EN-SL} nor Verifinger were able to correctly classify that pair.}
\label{fig:misclassified}
\end{center}
\vspace{-1.5em}
\end{figure*}

\par All the five individual models in the ensemble ($DP_o$, $DP_x$, $DP_y$, $DP_r$, and $DP_m$) comparable accuracy to each other. However, none of the individual models can match the accuracy of Verifinger, which underlines the limitations of relying on a single representation. The ensemble learning models based on decision fusion ($DP^{EN-DF}$) and feature fusion through external supervision  ($DP^{EN-SL}$) consistently outperform all the individual models and significantly close the gap to Verifinger accuracy. While decision fusion usually leads to marginally higher gains in performance compared to feature fusion, this improvement comes at the cost of increased computational requirements. The performance of the externally supervised model $DP^{EN-SL}$ is almost comparable with $DP^{EN-DF}$ while being much faster - in this case, almost five times faster since only one inference per image is required as opposed to five in decision fusion. Additionally, memory consumption is 5 times better in the case of $DP^{EN-SL}$ since no perturbations of the gallery are required as opposed to decision level fusion.
% \vspace{-1em}
% \begin{figure}
%     \centering
%     \includegraphics[scale=0.6]{images/failure_case.png}
%     \caption{Image illustrating an example where one model rejects a genuine pair of fingerprints as an impostor but another model correctly classifies it as genuine.  On the left are the original, unperturbed images evaluated using $M_{S}$ and on the right are the images flipped along the y-axis compared using $M_{Sy}$.  }
%     \label{fig:failure_case}
% \end{figure}
%\par We can see that the performance of the individual models is high and comparable. The reason we see an overall lower performance on the FVC database is largely due to the intentional distortion induced during the data collection. Since there are no latent fingerprints included in the training set, the accuracy on NIST SD27 and N2N are low. However, we can still observe the improvements as a result of ensemble learning in table \ref{tab:ensemble_one_to_one} which compares the $TAR \%$ at $FAR = 0.01\%$ of $DP^*$ along with the accuracies from decision-level, score-level fusion and feature-supervision when evaluated on all five databases. We can see that ... With the implementation of ensemble learning in fingerprint recognition, the output fusion techniques described in section \ref{sec:output_fusion} are implemented. 

\par Figure \ref{fig:misclassified} shows a few examples from NIST SD4, where the $DP_o$ model produces non-match errors (failure to match two fingerprints from the same finger). Of these 30 failure cases, $15$ of them can be rectified using the $DP^{EN-SL}$ model and this explains the better accuracy for the ensemble model. In addition to these 15 cases, the decision fusion approach $DP^{EN-DF}$ is able further correct two more errors. In comparison, Verifinger fails in only three of these 30 cases. This provides clear evidence that the single embedding generated by $DP^{EN-SL}$ is more diverse than the embedding generated by $DP_o$, thereby validating the proposed ensemble learning approach. However, the results also indicate that there is still some way to go to reach the accuracy levels of the COTS matcher.  

\subsection{Identification}

A large gallery of 1.2 million unique fingerprints is used in our evaluation \cite{yoon2015longitudinal}. For open-set identification, we use 1,000 non-mated and 1,000 mated fingerprint images from NIST SD4 as probes. For closed-set identification, we use 2,000 mated fingerprint images from NIST SD4 as probes. We repeat this procedure for NIST SD14 with the last 2,700 pairs. Additionally, we report identification results for two latent databases. While conducting open-set identification on the latent databases, half of the total number of rolled mates are included in the gallery. 

\begin{figure}[h]
    \centering
    \includegraphics[scale=0.4]{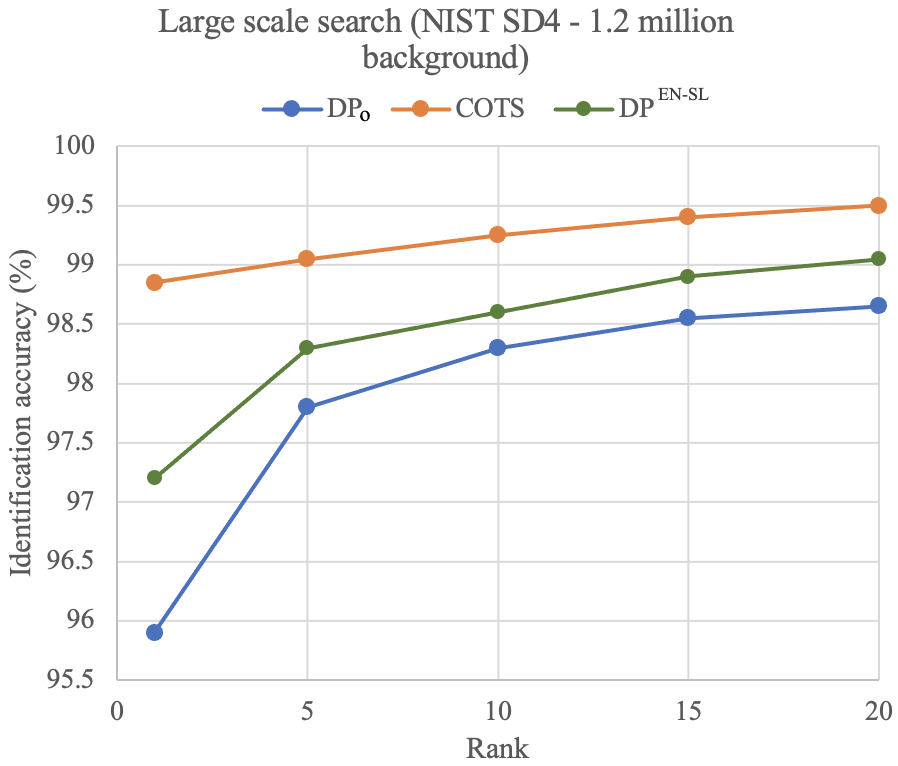}
    \caption{Cumulative match characteristic (CMC) for closed-set identification. Gallery size is 1.2 million images and the number of mated probes is 2,000.}
    \label{fig:cmc}
\end{figure}

Table \ref{tab:ensemble_closed_set} and Figure \ref{fig:cmc} show the results for closed-set identification. For decision fusion in closed-set identification, a search is deemed to result in a correct rank-1 retrieval if the probe's correct mate in the gallery is included in the rank-1 result of at least one of the component models in the ensemble. Clearly, both the decision ($DP^{EN-DF}$) and feature ($DP^{EN-SL}$) fusion ensemble models significantly outperform the best individual model in the ensemble ($DP_o$) on all the datasets. $DP^{EN-SL}$ performs at par with $DP^{EN-DF}$ at rank-1, while being five times faster than $DP^{EN-DF}$. Table \ref{tab:open_set} shows the results for open-set identification. Note that decision fusion schemes cannot be applied in the open-set scenario because of the presence of non-mated probes. This is a drawback of the decision fusion approach, which can be overcome using the $DP^{EN-SL}$ model that provides significant reduction in the false positive identification rate ($FPIR$) compared to the baseline model $DP_o$.

\begin{table}[h]
    \centering
    % \captionsetup{justification=centering}
    \caption{$FPIR(\%)$ at $FNIR = 0.3\%$ for open-set identification. Gallery size is 1.2 million images.}
        \begin{tabular}{|c | c | C{0.1\linewidth} | C{0.1\linewidth}|} 
             \hline
             Model & NIST SD4 & NIST SD14 & NIST SD27\\ [1.0ex]
             \hline
                $DP_o$ & 0.4 & 0.6 & 2.3\\
             \hline
                $DP^{EN-SL}$ & 0.2 & 0.44 & 1.5 \\ 
             \hline
                Verifinger & 0.1 & 0.19 & 0.7\\
             \hline
        \end{tabular}
    \label{tab:open_set}
\end{table}

\begin{table}[h]
    \centering
    % \captionsetup{justification=centering}
    \caption{Computational requirements for the various models evaluated on Intel(R) Core(TM) i7-8700K CPU @ 3.70GHz. Verifinger results are reported by the vendor and may depend on the number of minutiae/image.}
    \begin{threeparttable}
        \begin{tabular}{|c | C{0.3\linewidth}|} 
             \hline
             Model & Number of matches/second\\ [1.0ex]
             \hline
                $DP_o$            & $\sim$ 1M     \\
                \hline
                $DP^{EN-DF}$& $\sim$ 200K\\
                \hline
                $DP^{EN-SF}$  & $\sim$ 200K \\
                \hline
                $DP^{EN-SL}$ & $\sim$ 1M   \\
             \hline
                Verifinger & $\sim$ 15K-55K\\
            \hline
        \end{tabular}
        \end{threeparttable}
    \label{tab:computational_requirements}
\end{table}
% \vspace{-1.5em}

% \begin{table*}[ht]
%     \centering
%     \captionsetup{justification=centering}
%     \caption{Rank-1 accuracy for closed-set identification. Gallery size = 1.2 million and number of mated probes = 2,000.}
%     \begin{threeparttable}
%         \begin{tabular}{|c|c|c|c|c|c|}
%         \hline
%              \textbf{Database} & $DP_o$  & $DP^{EN-DF}$ & $DP^{EN-SF}$ & $DP^{EN-SL}$ & \textbf{Verifinger } \\
%              \hline
%              NIST SD4 & 95.9 & 97.1 & 96.2 & 97.1 & 98.9\\
%              \hline
%              NIST SD14 & 95.81 & 96.88  & 95.85 &  96.85 & 99.51\\
%              \hline
%             %  SD 27 & 25.19 & 27.9  & 25.99 &  & 26.7 & 27.13\\
%             %  \hline
%              NIST SD27 & 36.04 & 40.7  & 36.82 & 39.53 & 59.3\\
%              \hline    
%              N2N-RP & 71.2 & 74.7  & 72.0 & 73.9 & 82.1\\
%              \hline
%              N2N-RL & fill & Fill & Fill & Fill & Fill\\
%              \hline
%             %  FVC 2004 DB 2A & fill & fill  & fill & fill & Fill & Fill\\
%             %  \hline
%         \end{tabular}
%     \end{threeparttable}
%     \label{tab:ensemble_closed_set}
% \end{table*}

\begin{table}[ht]
    \centering
    \captionsetup{justification=centering}
    \caption{Rank-1 accuracy for closed-set identification on four databases. Gallery size = 1.2 million.}
    \begin{threeparttable}
        \begin{tabular}{|c|c|c|c|c|}
        \hline
             \textbf{Model} & SD4  & SD14 & SD27 & N2N\\
             \hline
             $DP_o$ & 95.9 & 95.81 & 14.34 & 71.2\\
             \hline
             $DP^{EN-DF}$ & 97.1 & 96.88  & 17.05 &  74.7 \\
             \hline
            %  SD 27 & 25.19 & 27.9  & 25.99 &  & 26.7 & 27.13\\
            %  \hline
             $DP^{EN-SF}$ & 96.2 & 95.85  & 15.5 & 72.0\\
             \hline    
             $DP^{EN-SL}$ & 97.1 & 96.85  & 15.9 & 73.9 \\
             \hline
             Verifinger & 98.9 & 99.51 & 42.24 & 82.1 \\
             \hline
            %  N2N-RL & fill & Fill & Fill & Fill & Fill\\
            %  \hlin
            %  FVC 2004 DB 2A & fill & fill  & fill & fill & Fill & Fill\\
            %  \hline
        \end{tabular}
    \end{threeparttable}
    \label{tab:ensemble_closed_set}
\end{table}

\subsection{Computational requirements}
% \vspace{-0.2em}
\par Using an ensemble of representations can have a significant impact on the retrieval times for search operations (see Table \ref{tab:computational_requirements}). This is where $DP^{EN-SL}$ demonstrates a critical advantage. Since there is no requirement to perturb either the probe or gallery image, the computational and memory requirements for $DP^{EN-SL}$ are the same as for $DP_o$. The search speed for $DP^{EN-SL}$ is almost five times faster than the search speed for $DP^{EN-DF}$ and $DP^{EN-SF}$. However, if memory and search speed is not a major concern, decision level fusion can be adopted while still being orders of magnitude faster than Verifinger. In practice, one also has to consider the memory required to store the training data for each model (may be needed in the future for updating/fine-tuning) in the ensemble, which linearly increases with the number of models in the ensemble.

% \begin{table}[]
%     \centering
%     \captionsetup{justification=centering}
%     \caption{Time taken to match a probe against the entire gallery (1.2 million)}
%         \begin{tabular}{|c | c|} 
%              \hline
%              Model & Time (ms) \\ [1.0ex]
%              \hline
%                 $M_{original}$            & 165    \\
%              \hline
%                 $M_{flip_{x}}$              & 150      \\
%              \hline
%                 $M_{flip_{y}}$              & 160      \\
%              \hline
%                 Ensemble & 375 \\ 
%              \hline
%         \end{tabular}
%     \label{tab:latency}
% \end{table}

\subsection{Ablation Study}
\label{sec:ablation_study}

\par We evaluate the contribution of each representation in the ensemble by performing 1:1 verification experiments on NIST SD4 using various subsets of representations from the ensemble. Table \ref{tab:ablation} summarizes the results of this study. While the two geometric transformations \textit{Flip\textsubscript{y}} and \textit{Flip\textsubscript{x}} are important for the decision fusion approach ($DP^{EN-DF}$), \textit{Ridge} and \textit{Minu} are critical for the success of feature fusion through external supervision. Simple feature concatenation did not lead to any accuracy improvement. Finally, to verify the stability of the results, we train the same network on the same input data starting with five different initializations and evaluate them on NIST SD4. Based on a t-test between the $TAR$ values of $DP_o$ and $DP^{EN-SL}$ over the five different initializations, the difference in the mean $TAR$ was found to be statistically significant at $0.05$ level.

\begin{table}[h]
    \centering
    \caption{Ablation study examining the individual contribution of each representation in the ensemble on NIST SD4. For reference, accuracy of baseline model $DP_o$ is $98.5\%$.}
    \begin{tabular}{|c|c|c|c|c}
    \hline
         \textbf{Subset $\mathcal{M}$} & $DP^{EN-DF}$ & $DP^{EN-SF}$ & $DP^{EN-SL}$ \\
         \hline
         $\{o,x\}$ & 98.75 & 99.0 & 98.95 \\
         \hline
         $\{o,m\}$ & 98.65 & 98.85 & 98.85  \\
         \hline
         $\{o,r\}$ & 98.65 & 98.95 & 98.95  \\
         \hline
         $\{o,y,x\}$ & 98.8 & 99.3 & 99.1  \\
         \hline
         $\{o,y,x,r,m\}$ & 98.75 & 99.3 & 99.0  \\
         \hline
         $\{o,r,m\}$ & 98.75 & 99.25 & \textbf{99.2}  \\
        %  \hline
        %  Binarized & 98.45 & 98.45 & 98.4 & 98.45 & 98.5 & 98.5\\
        %  \hline
        %  Gradient image & 98.45 & 98.45 & 98.45 & 98.45 & 98.45 & 98.45 \\
         \hline
    \end{tabular}
    \label{tab:ablation}
\end{table}

% \vspace{-1.5em}
\section{Summary}
\par In this work, we improve a SOTA deep-learning based fingerprint matcher (DeepPrint) by retraining it on manipulations of the original training dataset. We generated an ensemble of five fingerprint embeddings and proposed a feature fusion method that relies on external supervision of the individual representations to produce a more discriminative representation. This boosts the performance of the individual models without increasing computational requirements. We also considered decision and score fusion, which leads to further marginal improvement, albeit with higher computational complexity. These methods can serve as a wrapper that can be applied to any deep recognition system to boost overall performance.
{\small
\bibliography{egbib}
}

\end{document}